\documentclass[letterpaper, 10 pt, conference]{ieeeconf}  

\IEEEoverridecommandlockouts                              

\usepackage{graphicx}
\usepackage{subcaption}
\usepackage{verbatim}
\usepackage{fancyvrb}
\usepackage{framed}
\usepackage[hidelinks]{hyperref}
\usepackage{xcolor}
\usepackage{color}

\usepackage{float}   
\usepackage{color}
\usepackage[font=footnotesize]{caption}

\newif\ifblind
\blindfalse
\title{\LARGE \bf
HARMONIC: A Content-Centric Cognitive Robotic Architecture
}

\author{Sanjay Oruganti$^{1}$, Sergei Nirenburg$^{1}$, Marjorie McShane$^{1}$, Jesse English$^{1}$, Michael K. Roberts$^{1}$,\\ Christian Arndt$^{1}$, Carlos Gonzalez$^{3}$, Mingyo Seo$^{4}$, Luis Sentis$^{3}$
\thanks{$^{1}$Cognitive Science Department, Rensselaer Polytechnic Institute, Troy, NY 12180, USA. Departments of $^{3}$Aerospace Engineering and Engineering Mechanics, and ${^4}$ Electrical and Computer Engineering, at the University of Texas at Austin, TX, US.
        {\tt\small e-mail: sanjayovs@ieee.org}}%
}

\begin{document}


\maketitle
\thispagestyle{empty}
\pagestyle{empty}

\begin{abstract}
This paper introduces HARMONIC, a cognitive-robotic architecture designed for robots in human-robotic teams. HARMONIC supports semantic perception interpretation, human-like decision-making, and intentional language communication. It addresses the issues of safety and quality of results; aims to solve problems of data scarcity, explainability, and safety; and promotes transparency and trust. Two proof-of-concept HARMONIC-based robotic systems are demonstrated, each implemented in both a high-fidelity simulation environment and on physical robotic platforms. 

\end{abstract}


\section{Introduction.}
Recent advances in robotics have enabled robots to demonstrate impressive capabilities, from dynamic locomotion to dexterous manipulation \cite{tong2024advancements-0de}. However, impressive demonstrations often fail to translate into reliable real-world deployment and remain confined mainly to structured tasks such as sorting, navigation, and pick-and-place \cite{porter2024problems-394}. This disconnect arises because most robots perform well when executing learned policies \cite{yang2025boss-296} within their trained distributions, but struggle in long-horizon tasks in uncertain and unknown scenarios that characterize most real-world applications. 

Preparing a pizza is a good example of such a task for a robot. It involves a mixture of strategic (cognitive) and tactical (robotic control) operations. At a strategic level, a robot will have to perform compositional planning over extended action sequences (kneading dough, chopping vegetables); decompose high-level goals into executable sub-tasks (like preparing sauce); maintain causal dependencies across action chains (ensuring the oven reaches temperature before baking); execute under uncertainty (when certain ingredients are missing), etc. The above tasks require advanced reasoning, including causal inference and contingency planning across nested sub-plans with complex interdependencies, alongside robust execution under uncertainty, partial observability, and stochastic outcomes. At the tactical level, the robot must have precise sensorimotor control for physical execution, where learned policies must adapt to real-time feedback from force sensors, visual cues, and proprioceptive signals, while respecting safety constraints like collision avoidance and basic needs.

These strategic and tactical challenges have motivated recent work on integrating long-horizon planning and control through foundation model architectures that leverage different modalities and levels of abstraction \cite{kawaharazuka2024real-world-da9,bjorck2025gr00t}. In most, if not all, extant approaches, large Language Models (LLMs) serve as high-level semantic planners that decompose complex tasks into sub-goals and select appropriate skills from primitive action libraries \cite{raptis2025agentic-e5d}; while Vision-Language Models (VLMs) bridge the perception gap by jointly processing visual and textual inputs, enabling situated reasoning about physical scenes while maintaining semantic understanding \cite{zhang2025robridge-edf, ahn2024autort-938}.

\begin{figure}
    \centering
    \includegraphics[width=\linewidth]{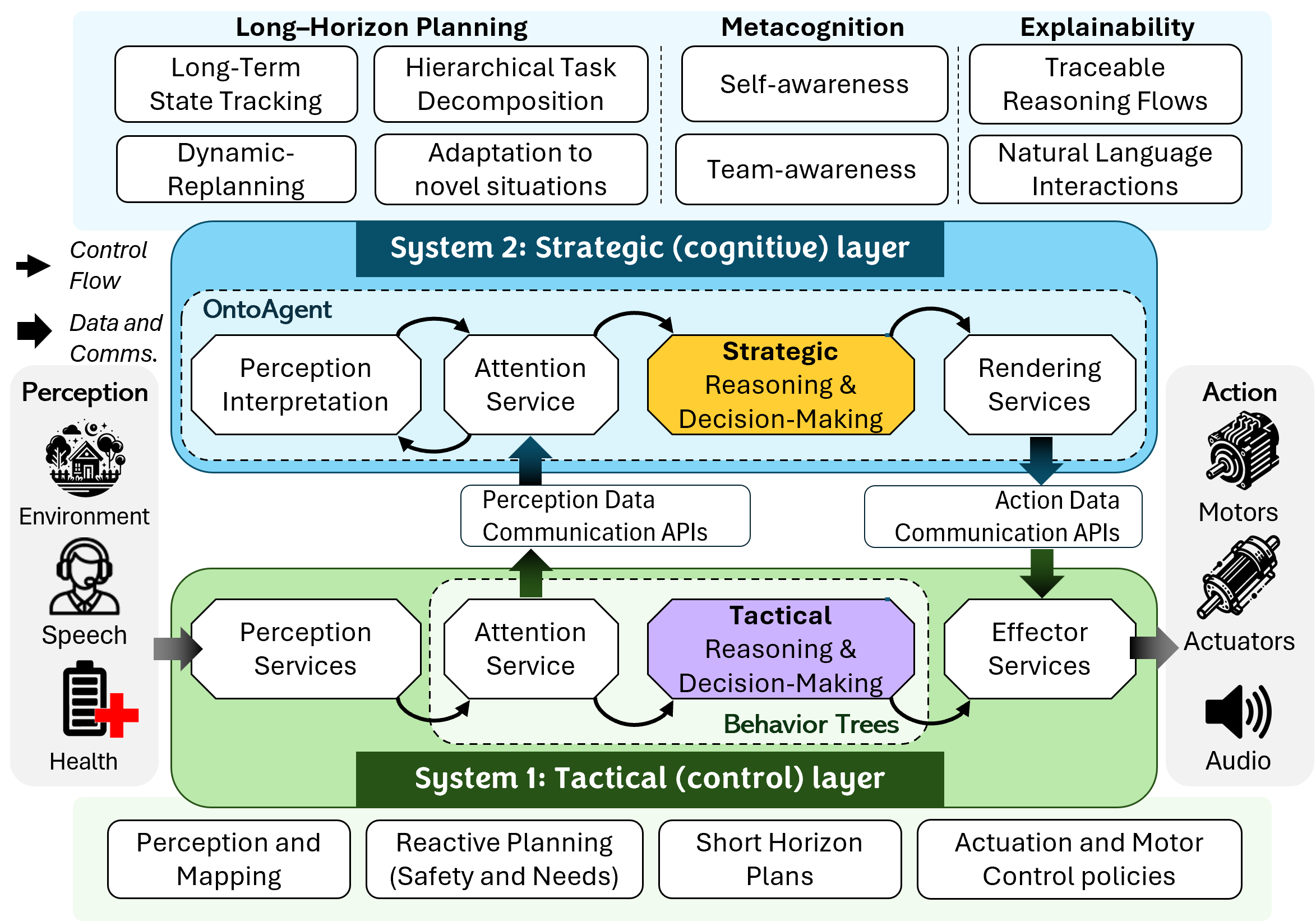}
    \caption{\footnotesize \textbf{Overview } of HARMONIC framework showing its Strategic and Tactical layers, corresponding to high-level deliberative planning (System 2) and low-level reactive control (System 1), and their respective capabilities.}
    \label{fig:overview}
    \vspace{-10pt}
\end{figure}

Vision Language Action models (VLAs) incorporate action decoders trained on robot demonstrations, directly outputting motor commands for both planning and control in a single model \cite{kawaharazuka2024real-world-da9}. Unlike LLM and VLMs, VLAs involve training on physical feasibility data, thus ensuring the feasibility and safety of robot action and obviating the need for affordance filters (such as those used, e.g., in SayCan \cite{ahn2022do-a14}). The latest state-of-the-art models include RT-2 \cite{zitkovich2023rt-2-889}, $\pi_0$ \cite{blacknoyearpi0-e8e}, and OpenVLA \cite{kim2024openvla-b4a}. Emerging hybrid architectures like GR00T \cite{bjorck2025gr00t} strategically combines VLMs for high-level planning with diffusion-based visuomotor policies for fast, reactive control.

While the above models demonstrate promising results in benchmark-based evaluations, a number of challenges remain -- data scarcity, cost of training, and issues with safety, quality of results, flexibility, explainability, transparency, and trust. 

Data scarcity and quality remain significant challenges for current systems. Training requires large-scale, diverse datasets across multiple modalities, necessitating specialized data acquisition strategies via teleoperation that are both expensive and difficult to scale. Model training is also computationally demanding because inputs and outputs are high-dimensional and multimodal. Fine-tuning pretrained models is more practical. Also, with the robot architectures continuing to evolve, constant data collection is unsustainable. To address this challenge, VLAs such as RT-2 \cite{zitkovich2023rt-2-889}, developed under the Open-X Embodiment initiative \cite{oneill2024open-808}, are being trained on data from a wide range of robotic embodiments. However, transferring policies across embodiments remains a major challenge. Separating strategic planning from tactical execution (as in SayCan \cite{ahn2024autort-938} and GR00T \cite{bjorck2025gr00t}) holds the promise of significantly reducing these costs.

The development of VLA models, such as RT-2 \cite{zitkovich2023rt-2-889}, illustrates a fundamental trade-off between generalization and efficiency. While these systems achieve strong multi-task generalization, the gains in motion quality and task success from larger architectures come at the cost of slower inference, higher latency, and greater computational demands. Hybrid approaches (e.g., GR00T \cite{bjorck2025gr00t}) attempt to balance this trade-off by enabling faster control execution while accommodating the slower timescales of high-level planning, supporting a case for a two-system approach.

In addition to data and compute demands, safety and reliability present steep challenges for these systems. Thus, hallucinations \cite{bai2024hallucination-eb2, raptis2025agentic-e5d} are a prominent issue across all modalities, arising from limitations in training data, architectural design, and a lack of context and understanding of the robots' working environment. More critically, in LLM- and VLM-based architectures, hallucinations also stem from a lack of knowledge of the consequences of actions or the execution of learned skills, leading to unsafe or unintended behaviors in real-world contexts. Unlike text-only models, compromised robotic agents can perform harmful physical actions, making conventional defenses such as filtering and action validation challenging and sometimes ineffective. It must be noted that hallucinations in these systems are no longer confined to text or image output errors but manifest as \textit{hallucinations in motion}, where misperceptions or faulty reasoning and control can translate directly into unsafe real-world behaviors. 

Beyond hallucination, these systems also exhibit significant concerns around bias, fairness, and transparency, with studies showing wide disparities in ethical safeguards, safety guardrails, and accountability measures \cite{sun2024comprehensive-6c4}. These limitations are further exacerbated in VLM and VLA systems, which multiply risks due to jailbreaking vulnerabilities as demonstrated by the recent ROBOPAIR attack framework \cite{robey2024jailbreaking-b9e}, inducing dangerous behaviors, including bomb detonation, covert surveillance, and human collisions. 

Cognitive-robotic architectures combine robotic control with cognitive supervision and offer a fundamentally different approach to the above challenges by grounding perception and action in verifiable knowledge structures, recognizing and communicating uncertainty instead of confidently producing incorrect interpretations \cite{mcshane2024agents-172}. They enable transparent, inspectable decision-making at all levels, also allowing for formal verification of safety constraints. Additionally, cognitive-robotic architectures allow incremental updating of their knowledge resources, eliminating the need to retrain models whenever new data, robots, or tasks are introduced. These architectures also achieve inference rates that are orders of magnitude faster than most foundation models.

In sum, this architectural approach directly addresses the limitations (briefly surveyed above) of current generative AI models. Most critically, the transparency inherent in cognitive-robotic systems enables meaningful human-robot collaboration, where team members can understand not just what the robot is doing but why, creating the accountability and trust essential for real-world deployment \cite{nirenburg2024explaining-315}. 

Long-horizon functioning has for decades been a core area of cognitive systems research \cite{nirenburg2018toward-d0e}. These systems excel at strategic reasoning -- extended attention management, goal selection, sophisticated planning capabilities, complex agent team interactions, etc. -- that current robotic approaches struggle to achieve. Historically, with few exceptions (see Related in Sec. \ref{sec:relatedwork}), cognitive systems have concentrated on applications that did not address the physical world and real-time demands of embodied agents. Integrating robotic systems that perform precise movements but fail when reasoning is required with cognitive systems, capable of transparent reasoning about complex tasks, is a natural next step in robotics. In this paper, we present our approach to this integration through the introduction of our novel HARMONIC framework.

\section{The HARMONIC Framework}
\label{sec:harmonic_framework}

Our framework, HARMONIC (Human-AI Robotic Team Member Operating with Natural Intelligence and Communication, Fig. \ref{fig:overview}), is an implemented dual-control cognitive robotic architecture featuring distinct layers of strategic reasoning and tactical, skill-level control \cite{oruganti2024harmonic-108}. This approach advances the hybrid control systems and architectures reviewed by Dennis et al. \cite{dennis2016practical} and contrasts with DIARC’s \cite{scheutz2013novel, schermerhorn2006diarc} integration strategy, which embeds the strategic layer within the tactical layer to support concurrent operation.

The strategic layer of HARMONIC adapts a mature cognitive architecture, OntoAgent \cite{english2020ontoagent-87e, mcshane2021linguistics-995, mcshane2024agents-172} for high-level reasoning, leveraging explicit, structured knowledge representations that can be inspected, verified, and incrementally expanded. (Space constraints allow us to present only a very limited description of OntoAgent in this paper; for a comprehensive account, see \cite{mcshane2024agents-172, mcshane2021linguistics-995}.) OntoAgent is built over a service-based ecosystem and includes, among others, processing modules for perception interpretation (a separate module for each perception modality), attention management, goal and plan selection, and plan execution. It employs both utility-based and analogical reasoning. Its knowledge substrate includes an ontological world model capable of supporting metacognition \cite{nirenburg2024mutual-5e0}, knowledge support for interpretation of perception, an episodic memory of past events, and a situation model that contains representations of entities and events that are part of the current task context. This layer prioritizes goals, manages plan agendas, and selects actions while continuously monitoring their execution. It also facilitates team-oriented operations, including communicating in natural language and explaining decisions. 

The tactical layer handles robotic control and execution, grounding cognitive capabilities in physical embodiment. This framework supports transparent, inspectable planning at both strategic and tactical levels: every decision, from high-level goal selection to individual motor commands, can be traced, inspected, understood, and formally verified. The framework grounds every perception and action in verifiable knowledge structures, recognizing and communicating uncertainty rather than confidently producing incorrect interpretations. 

The architecture also supports meaningful human-robot collaboration by allowing team members to understand not only what the robot is doing but also the reasoning behind its actions. HARMONIC-based robots generate and interpret explanations grounded in transparent causal reasoning, facilitating interventions from other team members in the robot's decision-making process. The above robotic capabilities are essential for real-world deployment where trust, safety, and accountability are paramount.

To support efficient inter-layer communication and data transfer, the strategic and tactical architectural layers of HARMONIC are connected through a bidirectional interface. Through this interface, the tactical layer provides the strategic layer with preprocessed multimodal (speech, vision, etc.) perception data and relays robot state information, employing a suite of perception models within the perception services. The strategic layer interprets this data, within the context of its situation model and active goals, and generates normalized ontologically grounded text (TMR) and vision (VMR) meaning representations that formally specify the semantics of perceptual input. The TMRs and VMRs are added to OntoAgent's situation model, which, together with ontology and other knowledge resources, provides data and heuristics for downstream functioning -- centrally, goal and plan agenda management, and execution. When plan execution reaches a step that corresponds to an atomic action, a command is issued for the tactical layer to execute it. 

The tactical layer, along with sensory input preprocessing, manages robot action planning, reflexive attention, and execution of motor actions in response to the high-level commands from the strategic layer. The attention service in this layer interprets perceptual inputs and assigns priority to sensory signals before passing them to the attention service in the strategic layer. Additionally, in the action translation pipeline of the tactical layer, specialized robotic planners, controllers, algorithms, and action sequence models are employed as part of the Reasoning and Decision Making modules and Effector Services to translate abstract action commands from OntoAgent into precise, executable robot control operations. 

\begin{figure}[h]
    \centering
    \includegraphics[width=\linewidth]{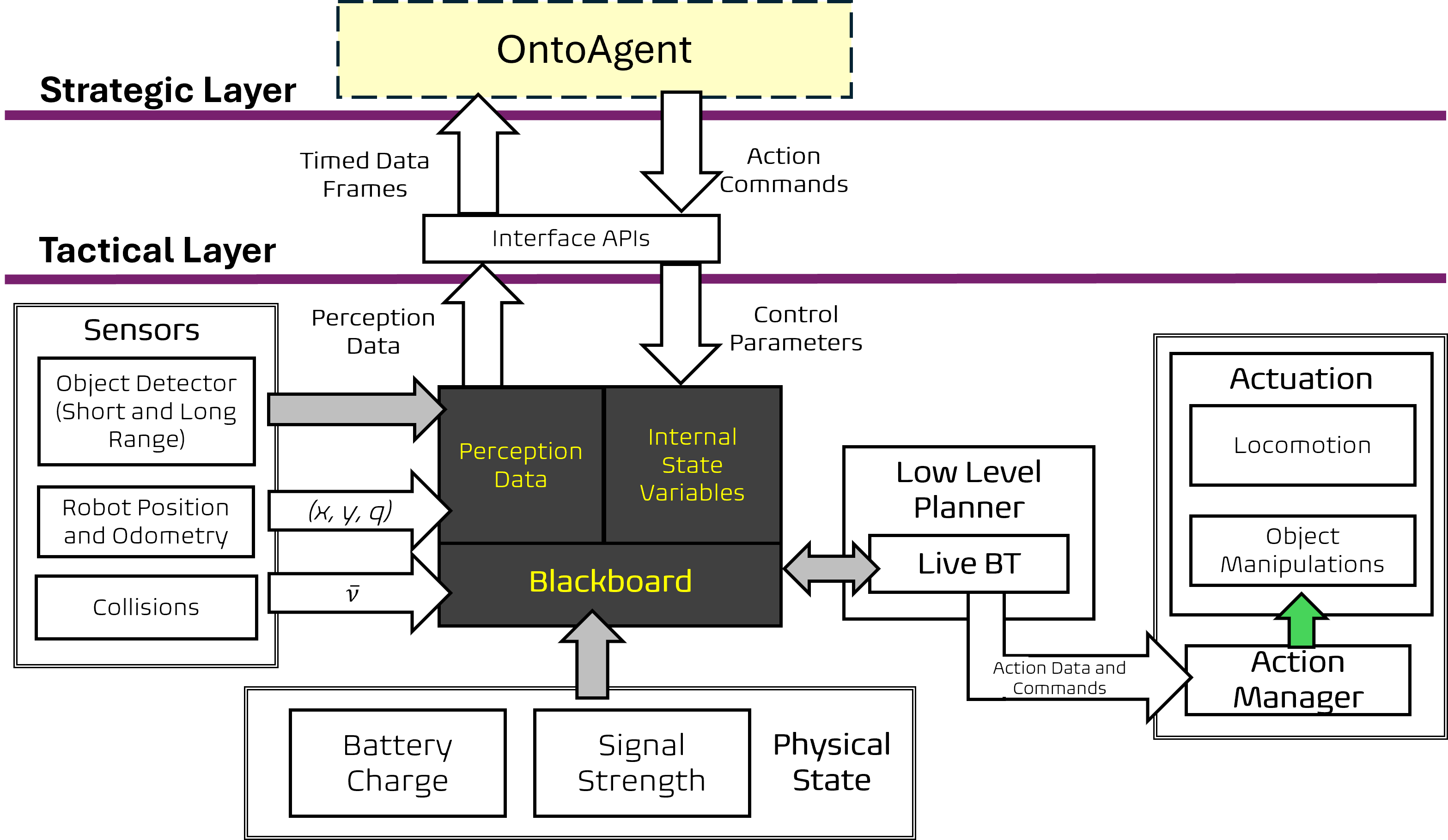}
    \caption{\footnotesize  Data-flow and control schematic illustrating the interactions between the behavior trees (BTs) in the tactical layer and the OntoAgent component in the strategic layer of the HARMONIC system.}
    \label{fig:Harmonic-sys}
\vspace{-5pt}
\end{figure}

When the OntoAgent issues a high-level command such as \texttt{DROP(key, at-position1)} (corresponding to "Drop the key on the ground"), the tactical planner decomposes this instruction into a coordinated sequence of operations of object recognition, spatial localization, trajectory planning, and motor execution. The tactical layer, implemented using Behavior Trees (BTs) \cite{colledanchise2018behavior}, provides dual functionality -- it executes the above sequence of operations, while maintaining a capability for dynamic environmental reactive responses. The hierarchical design structure of BTs enables real-time collision avoidance and adaptive behavior through embedded task prioritization mechanisms that can override or modify planned actions when necessary. The control models in BTs can selectively engage specialized policies, including whole-body compliant control, motion planning, and path planning algorithms \cite{iannotta2022heterogeneous, colledanchise2018behavior}. This modular architecture also facilitates the development of reusable skill primitives and transferable low-level plans that can be shared across different robotic platforms \cite{oruganti2025ikt1, venkata2023kt}.

The tactical planners use a blackboard to keep track of condition and state variables \cite{colledanchise2021implementation}, as shown in Fig. \ref{fig:Harmonic-sys}. This allows for efficient querying and updating of the system's state during operation based on sensory inputs and action commands received from the strategic layer. Sensory inputs are continuously recorded on the blackboard, and the attention services filter and package this perception data for transmission to the strategic layer. 

In parallel to this, the high-level action commands sent by the OntoAgent through the Interface APIs are unpacked and used to update the corresponding variables in the blackboard. This bidirectional flow ensures coherent state management while maintaining seamless coordination between perception processing and action execution. HARMONIC thus supports both real-time responsive behavior and deliberative higher-level planning capabilities, ensuring seamless coordination between perception processing and action execution.


The HARMONIC architecture reflects Kahneman’s dual-system framework \cite{kahneman2011thinking}, pairing reflexive and deliberative (System 1 and System 2) reasoning. The strategic layer is deliberative, enabling analytic decision making, while the tactical layer’s behavior trees (BTs) realize the reflexive, intuitive processes. By separating these modes of cognition, the architecture supports flexible scheduling and adaptive behavior and allows goals and actions to be reprioritized in real time. 

This dual-control arrangement equips HARMONIC to manage computational delays, unexpected contingencies, and safety constraints while optimizing resources through the use of low-level planning and reactive algorithms.

\section{Related Work}
\label{sec:relatedwork}
\vspace{-6pt}
Linking high-level reasoning with low-level sensorimotor control and implementation on embodied robots has been demonstrated using different strategies within well-known cognitive architectures, such as Soar \cite{laird2012cognitive}, ACT-R \cite{ritter2019act}, and DIARC \cite{scheutz2019overview}. 

Soar \cite{laird2012cognitive} utilizes multiple memory systems and the Spatial-Visual System (SVS) that bridges symbolic and continuous representations. While it excels at clear, structured decision-making and has been deployed on several robotic platforms from Puma arms \cite{laird1991robosoar} to REEM humanoids \cite{puigbo2015gpsr}, Soar requires extensive manual integration for perception and motor control. It faces the symbol grounding problem \cite{harnad1990symbol}, needing substantial customization to translate sensor data into symbols. Its serial decision cycle also limits real-time responsiveness in dynamic environments. 


The embodied extension of ACT-R/E \cite{trafton2013actre} enables human-robot interaction through wayfinding, grasping, and learning from demonstration. However, ACT-R's cognitive cycle calibrated to human speeds creates real-time constraints, while its single-threaded production system introduces computational overhead that limits suitability for time-critical robotic control.

DIARC \cite{scheutz2019overview,schermerhorn2006diarc} allows flexible instantiations with custom configurations tailored to specific platforms and tasks. It organizes perception, reasoning, and action modules into asynchronously operating parallel layers, featuring affect integration, language processing for situated dialogues \cite{scheutz2011humanlike}, and one-shot learning of new percepts, actions, and concepts from instruction. The architecture handles open-world uncertainty through its Goal Manager that coordinates behavior, while supporting task planning, incremental learning, and multi-agent communication through shared components. Despite these strengths, DIARC's distributed, flexible design introduces configuration complexity and debugging difficulties. At present, it lacks a realistic ontological model and English lexicon, long-term episodic memory, explicit metacognition, and formalized safety mechanisms, which limit its potential for real-world deployment. \cite{scheutz2019overview}.

These architectures underscore the inevitable need for fundamental trade-offs: Soar provides symbolic clarity but requires extensive customization for sensorimotor integration; ACT-R offers psychological fidelity but lacks the real-time responsiveness needed for dynamic environments; DIARC enables flexible distributed processing but sacrifices depth of understanding and systematic explainability. In developing HARMONIC, we faced similar trade-offs as well and decided to pay the price of having to acquire domain-specific knowledge in order for HARMONIC-based robots to operate with transparent intentions and understanding. To facilitate this process, we have developed protocols that accelerate the manual acquisition process and are actively developing tools for semi-automated knowledge acquisition with assistance from LLMs \cite{oruganti2023automating}. 

\begin{figure}[h]
     \centering
     \includegraphics[width=0.90\linewidth]{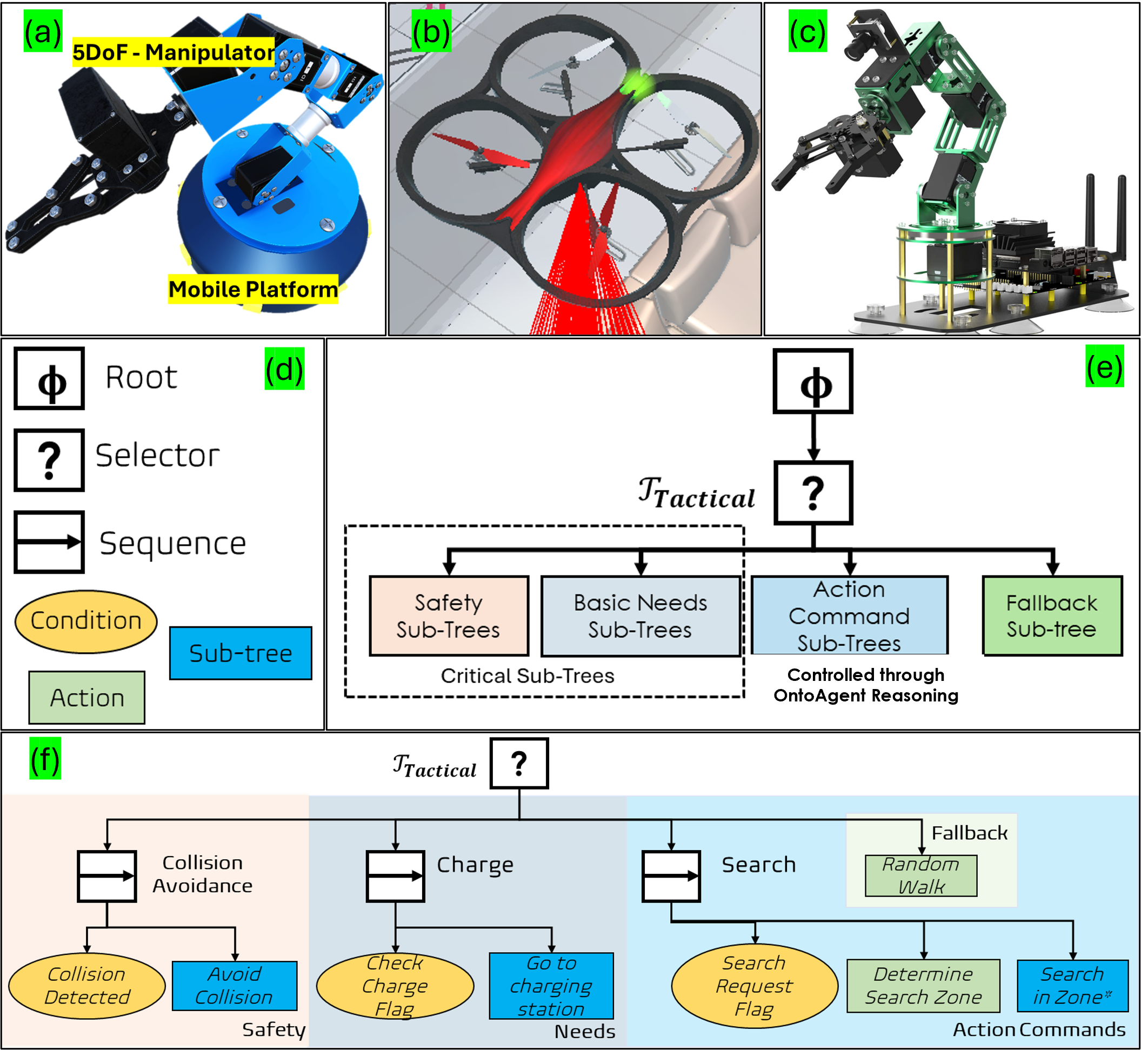}
     \caption{\footnotesize \textbf{(a)} UGV for ground-level exploration. \textbf{(b)} Parrot drone for aerial scanning. \textbf{(c)} 6-DoF robot. \textbf{(d)} Nodes in BTs. \textbf{(e)} BT design template for the robots’ tactical layer. \textbf{(f)} Sample BT on the UGV and 6-DoF robot.}
     \label{fig:bts_robots}
     \vspace{-10pt}
 \end{figure}

\section{Evaluation}
\label{sec:evaluation}
To evaluate HARMONIC, we implemented two proof-of-concept systems featuring HARMONIC-based robots in both simulation and hardware. One of the systems implements a robotic member of a shipboard maintenance team. The other system implements two robots performing a search-and-retrieve task under the guidance of a human. In the discussion below, we describe and demonstrate the functioning of these systems. First, we present the robots and their tactical capabilities. Next, provide the setup details for the scenarios in which each of the systems operates.

\subsection{Robot Platforms and Tactical Capabilities}

Across the two systems, we use three different robots: a UGV and a Drone in simulation environments, and a tabletop serial manipulator. Each robot runs on HARMONIC: it operates its own instance of OntoAgent at the strategic level and customized low-level planners at the tactical level. In addition to general knowledge, each of the robots has personalized knowledge, e.g., ontological scripts detailing complex events in which they are expected to participate or which they are expected to comprehend when performed by teammates. This knowledge circumscribes their team roles and capabilities. Plans that robots generate and carry out are instances of scripts with parameters adjusted for the robot's context. These plans are created by running one or more ontologically stored meta-scripts -- procedural knowledge that HARMONIC robots use to consciously make decisions about their understanding, scheduling, and action.\footnote{All HARMONIC robots operate this way, but due to space limits we detail the knowledge and planning organization only for the shipboard maintenance system and provide more information on the project page \href{https://tinyurl.com/hicra26}{tinyurl.com/hicra26}. A full description of OntoAgent and its resources appears in \cite{mcshane2021linguistics-995, mcshane2024agents-172}, and the attached video presents real-time evaluations.}


\begin{figure*}[ht]
    \centering
    \includegraphics[width=\linewidth]{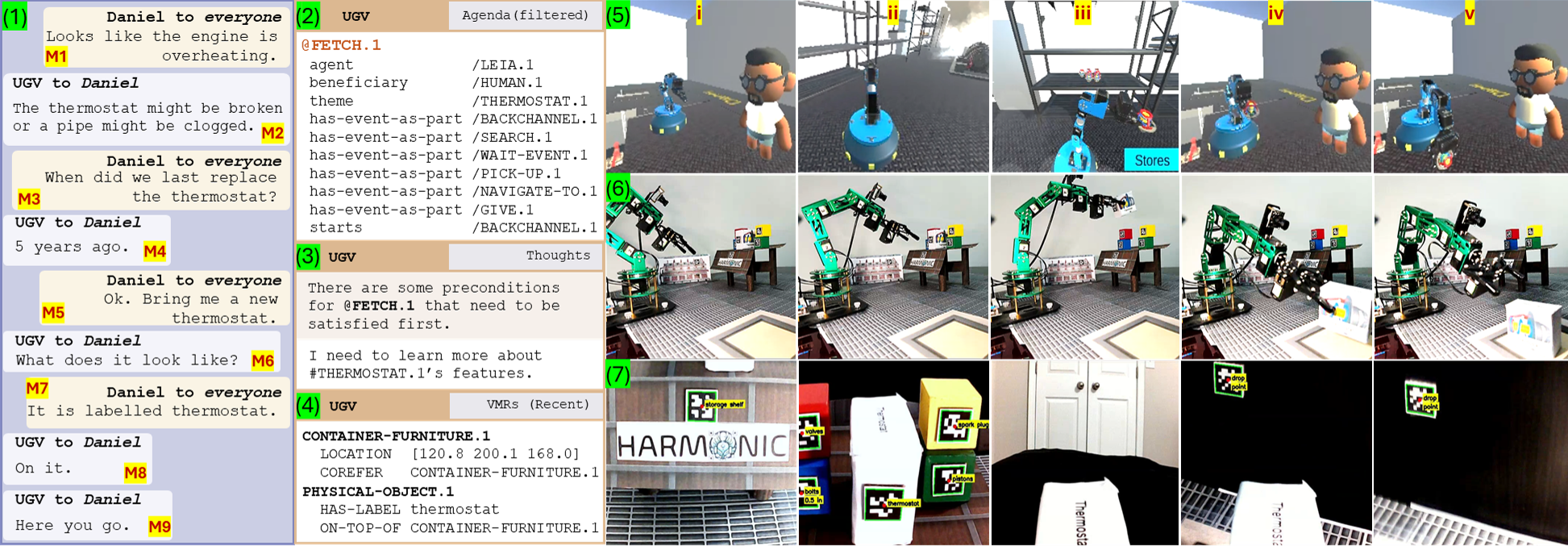}
    \caption{\footnotesize\textbf{(1)-(4)} Are the panels from DEKADE UI. \textbf{(1)} Communication transcript between Daniel (human) and the robots. \textbf{(2)} Robot's task agenda. \textbf{(3)} Complete reasoning transcripts of the robot. \textbf{(4)} Sample Vision Meaning Representations (VMRs) of detected objects from the UGV (leader). \textbf{(5)} Simulated ship environment scenes displaying the UGV. \textbf{(6)} Tabletop robot performing various tasks. \textbf{(7)} FPV camera view from the manipulator. Robot task snapshots: \textbf{(i)} Initial position. \textbf{(ii)} Searching stores for thermostat. \textbf{(iii)} Picking up thermostat. \textbf{(iv)} Returning thermostat to Daniel. \textbf{(v)} Dropping thermostat at location.}
    \label{fig:Evaluation_Results_Robot_thermostat}
    \vspace{-15pt}
\end{figure*}

At the tactical layer, BTs offer a flexible and intuitive approach for designing control actions and establishing task-planning hierarchies for robots. BTs handle tactical-level attention management, implement tactical static plans, and activate both policies and model-based controllers through selective engagement. All controllers are maintained in the action manager depicted in Fig. \ref{fig:Harmonic-sys}. In all the robots we use, BTs follow a standardized design template that prioritizes safety and operational requirements as shown in Fig. \ref{fig:bts_robots}. This design uses the conventional left-to-right execution order, ensuring the robot's critical needs and priorities are addressed. Collision avoidance subtrees are positioned on the left, followed by robot-needs subtrees, then action-command subtrees. On the right side, fallback subtrees handle behaviors such as random exploration. An example from the UGV's BT is presented in Fig. \ref{fig:bts_robots}.

The integration infrastructure of HARMONIC for these robots is developed in ROS2 for both the robots and simulators, and the simulation environments are created in Unity. OntoAgent is developed in Python, incorporating various linguistic, machine learning, and statistical tools, the details of which are presented in \cite{mcshane2021linguistics-995, mcshane2024agents-172}. We plan to release the HARMONIC infrastructure with appropriate licensing for research purposes in the near future.

\subsection{The Shipboard Maintenance System}
\label{sec:shipboard_maintenance}
\textbf{Scenario:} \textit{In a shipboard maintenance team, the robot assistant,  LEIA\footnote{We use the name LEIA interchangeably with OntoAgent. LEIA stands for Language Endowed Intelligent Agents}, interacts conversationally with the maintenance mechanic, Daniel, assists in diagnosing engine issues, and supports maintenance tasks by locating and retrieving necessary replacement parts.}  

We demonstrate this scenario a) in a simulated ship environment using a UGV and b) with a tabletop robot that performs the tasks in a tabletop setup. The verbal interactions (\textbf{M1-M9}) in this scenario are shown in Fig. \ref{fig:Evaluation_Results_Robot_thermostat} and in the accompanying video. The figure also shows snapshots of the robot's agenda, a rendition in English of its internal thoughts, and current VMRs. The TMRs for a subset of the utterances in the scenario are described below.



\subsubsection{Task Initiation} Daniel utters \textbf{(M1)}. LEIA interprets this input and represents it as the following TMR:

\begin{codeblock}
\textbf{#DESCRIBE-MECHANICAL-PROBLEM.1}
 agent       #HUMAN.1    //Speaker
 beneficiary #LEIA.1     //Robot
 theme       #OVERHEAT.1 //Overheating issue
\textbf{#OVERHEAT.1}
 theme      #ENGINE.1  //Engine is what is overheating
\textbf{#ENGINE.1}
 corefer   ->ENGINE.1  //That specific engine in the room
\end{codeblock}
The TMR contains an instance of the ontological concept \texttt{DESCRIBE-MECHANICAL-PROBLEM} whose \texttt{THEME} is an instance of the concept \texttt{OVERHEAT} whose \texttt{THEME}, in turn, is an instance of the concept \texttt{ENGINE}. There is only one engine in the scene, so there is no need for reference resolution.\footnote{Once again, language interpretation in OntoAgent cannot be detailed here due to space constraints. Please visit \href{https://tinyurl.com/hicra26}{tinyurl.com/hicra26} for details.} The robot knows that overheating is a sign of malfunction. It is also capable of inferring that Daniel's (and, therefore, the team's) goal is to resolve this problem. It also knows that to resolve problems, they must be first diagnosed and that its role as an assistant involves helping Daniel to diagnose the problem. As a result, the robot puts the goal of diagnosing the problem on its agenda. Ontological specifications of robots' goals include pointers to associated scripts (plan templates). The above goal is associated with a single script, and LEIA creates an instance of it, a plan, as follows:
\begin{codeblock}
\textbf{Plan.1}
 \textbf{#HYPOTHESIZE-MECHANICAL-PROBLEM-CAUSE.1}
   agent       #LEIA.1     //The agent will respond
   beneficiary #HUMAN.1    //to the speaker
   theme       #OVERHEAT.1 //about engine's temperature
   *take-this-action "search ontology for causes;
                      report."
\end{codeblock}

\subsubsection{Problem Identification} \texttt{Plan.1} does not involve preconditions. So, LEIA executes its first step, the ontology search procedure, and finds related causes: \texttt{OBSTRUCT} (possible pipe obstruction), \texttt {STATE-OF-REPAIR} (relating to a possible thermostat issue). Next, it executes the report step of the plan. LEIA first generates a representation of the content to be generated (GMR) and then uses its native semantic text generator \cite{mcshane2021linguistics-995,mcshane2024agents-172} to generate the English utterance conveying that content (\textbf{M2}). 

\begin{codeblock}
\textbf{#ALTERNATIVE.1}      //It might be either of two options
 domain #MODALITY.1
 range  #MODALITY.2
\textbf{#MODALITY.1}         //that a pipe is obstructed
 type   EPISTEMIC
 value  0.5
 scope  #OBSTRUCT.1
\textbf{#MODALITY.2}         //or the thermostat is broken
 type   EPISTEMIC
 value  0.5
 scope  #STATE-OF-REPAIR.1
\textbf{#OBSTRUCT.1}
 theme  @PIPE
\textbf{#STATE-OF-REPAIR.1}
 domain @THERMOSTAT
 range  <0.7 \hfill \textbf{\textcolor{blue}{The GMR for M2}}
\end{codeblock}

Unlike LLMs, which can generate high-quality text but lack understanding of their own processes and how they operate, the generator in HARMONIC is fully transparent, allowing it to explain its behavior. LLMs are, in fact, used at one stage of the generation process -- but only for the semantically vacuous task of selection of the best contextual option from several semantically correct and pragmatically appropriate candidate utterances. 

The processing triggered by \textbf{(M3)} and that yielding \textbf{(M4)} is similar to that described above. At this point, both Daniel and LEIA know that the thermostat is too old and should be replaced. Daniel immediately utters \textbf{(M5)}, which obviates the need for LEIA to generate a suggestion similar to that in \textbf{(M2)}.

\subsubsection{Information Gathering}  LEIA interprets the following TMR from \textbf{(M5)}:
\begin{codeblock}
\textbf{#REQUEST-ACTION-FETCH.1}
 agent       #HUMAN.1     //Speaker asks
 beneficiary #LEIA.1      //Listener to fetch
 theme       #THERMOSTAT.1//a thermostat
\textbf{#THERMOSTAT.1}
 age        0.0001<>0.1   //thats new.
\end{codeblock}

\begin{figure*}[ht]
    \centering
    \includegraphics[width=\linewidth]{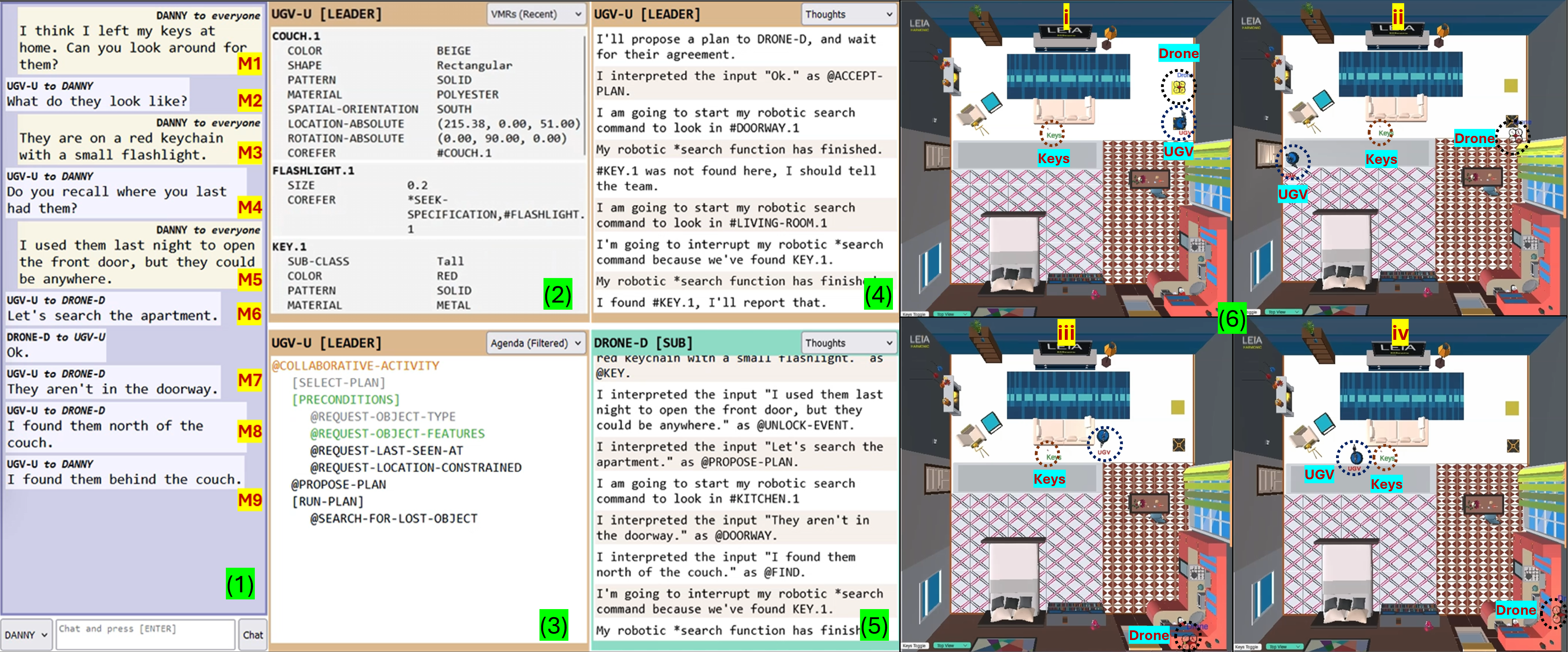}
    \caption{\footnotesize\textbf{(1)} Communication transcript between Daniel (human) and the robots. \textbf{(2)} Example VMRs of objects detected by UGV. \textbf{(3)} UGV's task agenda. \textbf{(4, 5)} Reasoning transcripts (thoughts) of UGV and Drone. \textbf{(6)} Snapshots from the simulations showing the UGV, drone, and keys. {\textbf{(i)}} Robots at their stations. {\textbf{(ii, iii)}} Robots conducting \texttt{SEARCH}. {\textbf{(iv)}} Keys located by the UGV}
    \label{fig:Demo2_Apartment_Search}
    \vspace{-15pt}
\end{figure*}

LEIA posits a new goal to carry out Daniel's command, generates a plan, \texttt{FETCH.1}, from a script associated with the goal, and puts it on the agenda as shown in Fig. \ref{fig:Evaluation_Results_Robot_thermostat}(2). When it attempts to execute the plan, it finds that some of the properties (features) of the thermostat in question are not known. Hence, the agent looks in its ontology for a meta-plan that can satisfy the precondition and finds a plan involving requesting information from a teammate, which results in generating a question, \textbf{(M6)}. Daniel responds with \textbf{(M7)}, and the agent backchannels with \textbf{(M8)}. 

\subsubsection{Fetch Execution} 
 The agent continues executing the \texttt{FETCH.1} plan by executing the \texttt{SEARCH}, \texttt{HOLD}, \texttt{RETURN}, and \texttt{DROP} tactical-level plans, as illustrated in Fig. \ref{fig:Evaluation_Results_Robot_thermostat}(5,6,7)i-v on both simulations and a tabletop robot. Each plan triggers a sequence of steps (atomic actions) with specific parameters. For example, \texttt{SEARCH} requires information about object features (this is why LEIA engaged in the above interaction with Daniel) and a location, which in this case is retrieved from its episodic memory. \texttt{HOLD} needs the object type and its features, \texttt{RETURN} uses Daniel's location as a waypoint parameter, and \texttt{DROP} uses the floor as a relative location.

\subsubsection{Task Completion} During \texttt{SEARCH}, when the thermostat is found and identified as such, the tactical controller stops and sends feature information to the strategic layer for matching against stored knowledge. The tactical-level search plan is designed to implement \textit{attribute-based search} strategy~\cite{kahneman2011thinking}, which involves substituting a difficult question (looking for a thermostat with a specific label) with a simpler one (looking for all thermostat-like objects first and then looking for a label). Once a thermostat-like object is found, the object feature data is sent to the strategic layer by first creating a VMR for this candidate object\footnote{The current implementation simplifies this step by labeling objects with QR-codes that link the object with its corresponding features.}, as shown in Fig.~\ref{fig:Evaluation_Results_Robot_thermostat}(4). The VMR is then grounded with the expected object features and confirmed. Once confirmed, the \texttt{PICKUP} action is executed, followed by \texttt{RETURN} and \texttt{DROP}.  LEIA  maintains natural language communication with Daniel throughout this process, as shown in \textbf{(M8, M9)}.

\subsection{Demo 2: Team Search}
\textbf{Scenario:} \textit{In an apartment environment, a heterogeneous multi-robot team consisting of a UGV and a drone assists a human, Danny, in locating a lost set of keys.}

We demonstrate this scenario in a simulated apartment environment shown in (Fig.~\ref{fig:Demo2_Apartment_Search}). Much of the processing in this scenario follows the template of the shipboard scenario. So, we will comment on those features that are different.  As mentioned above, each robot operates its own instance of OntoAgent and possesses knowledge of the team hierarchy. In the current set-up, the UGV is designated as the leader. As a result of this, its knowledge base is deeper than that of its subordinate, the drone. This is because it, unlike the drone, is responsible for setting goals, generating plans from scripts, resolving unmet preconditions, providing instructions to subordinates, and communicating with the human.
The robots know the layout of the apartment and know which areas of the apartment are easiest to access by each of them.



\subsubsection{Task Initiation}
The scenario begins with Daniel asking a question \textbf{(M1)}. Once this input is interpreted by the UGV, it puts the team-level goal of finding the keys on its agenda. The goal is associated with the metalevel script for \texttt{COLLABORATIVE-ACTIVITY} and several scripts, of which \texttt{SEARCH-FOR-LOST-OBJECT} is selected. Plans generated from these scripts are put on agenda. Meanwhile, the drone remains on standby awaiting instructions.

\subsubsection{Information Gathering}
Before the selected plan, \texttt{SEARCH-FOR-LOST-OBJECT} can be executed, its preconditions must be checked, as shown in Fig.~\ref{fig:Demo2_Apartment_Search}(3). The object type (keys) is already known, but the last-seen location and other descriptive features are not. The UGV queries Daniel for this information and receives responses (\textbf{(M2-M5)}. It can be noted that \textbf{(M4)} enables the robot to prioritize the \textit{entryway} in the search sequence \textbf{(ii)}.

\subsubsection{Search Execution}
The UGV instructs the drone to initiate the search \textbf{(M6)}, and both robots begin exploring their preassigned areas as shown in \textbf{(ii–iv)}. The tactical modules on each robot control the search process using the waypoint strategy, while the strategic (cognitive) module maintains awareness of area coverage without directly guiding robot navigation.

\subsubsection{Communication and Coordination}
Throughout the search, the robots communicate their findings to each other and to the human \textbf{(M7–8, ii–iv)}. When the keys are not found in a certain area, the information is communicated to the team.

\subsubsection{Task Completion}
When a robot (in this case, the UGV) finds the keys, its attention module commands it to halt and check the keys' features against the internal representation of the keys that was completed during the information gathering step -- checking that what it sees are indeed Danny's keys. The features align, so the search is aborted, and the robot that found the keys (in this case, the UGV) issues a corresponding notification to the robotic teammate \textbf{(M8)} and communicates to Danny using an utterance \textbf{(M9)}, more suitable for a human teammate.
\vspace{-10pt}
\section{Discussion}
\label{sec:discussion}
We demonstrated HARMONIC’s long-horizon planning capabilities by illustrating how LEIA a) dynamically assembles and revises plan sequences in response to situational changes, b) uses metaplanning to anticipate precondition satisfaction needs and recover from knowledge lacunae. At the tactical level, reactive planning integrates smoothly with skill execution, assuring safety. Thus, the maintenance robot seamlessly transitions from search to manipulation to delivery while managing collision avoidance. The clear separation of strategic and tactical layers enables seamless cross-embodiment adaptation and sim-to-real transfer. OntoAgent's best-in-class language understanding and generation capabilities enable the robots to interpret complex directives and engage in intentional, contextually grounded, human-level dialog.

 While VLA-based architectures may claim multiple abstraction levels through automatic scaling, these remain opaque and uninspectable. Similarly, current LLM/VLM embodiment strategies focus on language-to-action mapping without addressing the affordance and actionability knowledge gaps or the ever-present issue of hallucinations. HARMONIC does not suffer from the above constraints. Its integration of a cognitive architecture with robotic control is theoretically motivated by Tversky and Kahneman's dual-system framework \cite{kahneman2011thinking} and engineered using sophisticated APIs and a well-organized, transparent knowledge substrate supporting both the strategic and tactical processing.

 The core current limitation of the current approach is the need to grow the knowledge resources at both the strategic and tactical levels of HARMONIC. We are well aware of this issue's importance and are pursuing this direction of work along several avenues. We have developed and put to use LLM-enhanced support tools \cite{oruganti2023automating} and training materials for human acquirers \cite{mcshane2024agents-172}(Ch.9). More importantly, we are actively developing agent capabilities to learn knowledge automatically through instruction, demonstration, and guidance from their human teammates \cite{mcshane2024agents-172}(Ch.7), \cite{nirenburg2023hybrid,nirenburg2021overcoming}. 

In the present study, we focus on qualitative evaluation because few existing systems are directly comparable to ours, making benchmarking or controlled comparison experiments infeasible. However, we plan to extend this work with targeted comparative studies against foundation models, examining factors such as hallucination rates, efficiency, and reliability. We also plan to align HARMONIC with the Agentic AI movement by introducing OntoAgentic AI, with OntoAgent serving as the orchestrator instead of an LLM.





\ifblind
\else
\section*{ACKNOWLEDGMENT}
This work was supported in part by ONR Grant \#N00014-23-1-2060. The views expressed are those of the authors and do not necessarily reflect those of the Office of Naval Research.
\fi


\bibliographystyle{IEEEtran}
\bibliography{references,references_2}

\begin{thebibliography}{10}
\providecommand{\url}[1]{#1}
\csname url@samestyle\endcsname
\providecommand{\newblock}{\relax}
\providecommand{\bibinfo}[2]{#2}
\providecommand{\BIBentrySTDinterwordspacing}{\spaceskip=0pt\relax}
\providecommand{\BIBentryALTinterwordstretchfactor}{4}
\providecommand{\BIBentryALTinterwordspacing}{\spaceskip=\fontdimen2\font plus
\BIBentryALTinterwordstretchfactor\fontdimen3\font minus \fontdimen4\font\relax}
\providecommand{\BIBforeignlanguage}[2]{{%
\expandafter\ifx\csname l@#1\endcsname\relax
\typeout{** WARNING: IEEEtran.bst: No hyphenation pattern has been}%
\typeout{** loaded for the language `#1'. Using the pattern for}%
\typeout{** the default language instead.}%
\else
\language=\csname l@#1\endcsname
\fi
#2}}
\providecommand{\BIBdecl}{\relax}
\BIBdecl

\bibitem{tong2024advancements-0de}
Y.~Tong, H.~Liu, and Z.~Zhang, ``Advancements in humanoid robots: A comprehensive review and future prospects,'' \emph{{IEEE}/{CAA} Journal of Automatica Sinica}, vol.~11, no.~2, pp. 301--328, 2024.

\bibitem{porter2024problems-394}
\BIBentryALTinterwordspacing
B.~Porter, ``The problems with humanoid robots,'' 5 2024. [Online]. Available: \url{https://medium.com/@bp_64302/the-problems-with-humanoid-robots-9d8684d62008}
\BIBentrySTDinterwordspacing

\bibitem{yang2025boss-296}
Y.~Yang, L.~Zhao, M.~Ding, G.~Bertasius, and D.~Szafir, ``{BOSS}: Benchmark for observation space shift in long-horizon task,'' \emph{{IEEE} Robotics and Automation Letters}, vol.~10, no.~9, pp. 8882--8889, 2025.

\bibitem{kawaharazuka2024real-world-da9}
K.~Kawaharazuka, T.~Matsushima, A.~Gambardella, J.~Guo, C.~Paxton, and A.~Zeng, ``Real-world robot applications of foundation models: a review,'' \emph{Advanced Robotics}, vol.~38, no.~18, pp. 1232--1254, 2024.

\bibitem{bjorck2025gr00t}
J.~Bjorck, F.~Casta{\~n}eda, N.~Cherniadev, X.~Da, R.~Ding, L.~Fan, Y.~Fang, D.~Fox, F.~Hu, S.~Huang \emph{et~al.}, ``Gr00t n1: An open foundation model for generalist humanoid robots,'' \emph{arXiv preprint arXiv:2503.14734}, 2025.

\bibitem{raptis2025agentic-e5d}
E.~K. Raptis, A.~C. Kapoutsis, and E.~B. Kosmatopoulos, ``Agentic {LLM}-based robotic systems for real-world applications: a review on their agenticness and ethics,'' \emph{Frontiers in Robotics and {AI}}, vol.~12, p. 1605405, 2025.

\bibitem{zhang2025robridge-edf}
K.~Zhang, R.~Xu, P.~Ren, J.~Lin, H.~Wu, L.~Lin, and X.~Liang, ``{RoBridge}: A hierarchical architecture bridging cognition and execution for general robotic manipulation,'' \emph{{arXiv}}, 2025.

\bibitem{ahn2024autort-938}
M.~Ahn, D.~Dwibedi, C.~Finn, M.~G. Arenas, K.~Gopalakrishnan, K.~Hausman, and e.~al., ``{AutoRT}: Embodied foundation models for large scale orchestration of robotic agents,'' \emph{{arXiv}}, 2024.

\bibitem{ahn2022do-a14}
M.~Ahn, A.~Brohan, N.~Brown, Y.~Chebotar, O.~Cortes, B.~David, and e.~al., ``Do as i can, not as i say: Grounding language in robotic affordances,'' \emph{{arXiv}}, 2022.

\bibitem{zitkovich2023rt-2-889}
B.~Zitkovich, T.~Yu, S.~Xu, P.~Xu, T.~Xiao, F.~Xia, and e.~al., ``{RT}-2: Vision-language-action models transfer web knowledge to robotic control,'' in \emph{Proceedings of The 7th Conference on Robot Learning}, ser. Proceedings of Machine Learning Research.\hskip 1em plus 0.5em minus 0.4em\relax {PMLR}, 2023.

\bibitem{blacknoyearpi0-e8e}
K.~Black, N.~Brown, D.~Driess, A.~Esmail, M.~Equi, C.~Finn, and e.~al., ``\${\textbackslash} pi\$0: A vision-language-action flow model for general robot control.'' \emph{{arXiv} preprint {ARXIV}.2410.24164}.

\bibitem{kim2024openvla-b4a}
M.~J. Kim, K.~Pertsch, S.~Karamcheti, T.~Xiao, A.~Balakrishna, S.~Nair, and e.~al., ``{OpenVLA}: An open-source vision-language-action model,'' \emph{{arXiv}}, 2024.

\bibitem{oneill2024open-808}
A.~O’Neill, A.~Rehman, A.~Maddukuri, A.~Gupta, A.~Padalkar, A.~Lee, and e.~al., ``Open x-embodiment: Robotic learning datasets and {RT}-x models,'' \emph{2024 {IEEE} International Conference on Robotics and Automation ({ICRA})}, vol.~00, pp. 6892--6903, 2024.

\bibitem{bai2024hallucination-eb2}
Z.~Bai, P.~Wang, T.~Xiao, T.~He, Z.~Han, Z.~Zhang, and M.~Z. Shou, ``Hallucination of multimodal large language models: A survey,'' \emph{{arXiv}}, 2024.

\bibitem{sun2024comprehensive-6c4}
F.~Sun, R.~Chen, T.~Ji, Y.~Luo, H.~Zhou, and H.~Liu, ``A comprehensive survey on embodied intelligence: Advancements, challenges, and future perspectives,'' \emph{{CAAI} Artificial Intelligence Research}, p. 9150042, 2024.

\bibitem{robey2024jailbreaking-b9e}
A.~Robey, Z.~Ravichandran, V.~Kumar, H.~Hassani, and G.~J. Pappas, ``Jailbreaking {LLM}-controlled robots,'' \emph{{arXiv}}, 2024.

\bibitem{mcshane2024agents-172}
M.~{McShane}, S.~Nirenburg, and J.~English, \emph{Agents in the Long Game of {AI}: Computational Cognitive Modeling for Trustworthy, Hybrid {AI}}.\hskip 1em plus 0.5em minus 0.4em\relax Cambridge, {MA}: {MIT} Press, 2024.

\bibitem{nirenburg2024explaining-315}
S.~Nirenburg, M.~{McShane}, K.~Goodman, and S.~Oruganti, ``Explaining explaining,'' \emph{Proceedings of the 1st International Conference on Explainable {AI} for Neural and Symbolic Methods}, pp. 116--123, 2024.

\bibitem{nirenburg2018toward-d0e}
S.~Nirenburg, M.~{McShane}, S.~Beale, P.~Wood, B.~Scassellati, O.~Magnin, and A.~Roncone, ``Toward human-like robot learning,'' in \emph{International Conference on Applications of Natural Language to Information Systems}.\hskip 1em plus 0.5em minus 0.4em\relax Springer, 2018.

\bibitem{oruganti2024harmonic-108}
S.~Oruganti, S.~Nirenburg, M.~{McShane}, J.~English, M.~K. Roberts, and C.~Arndt, ``{HARMONIC}: A framework for explanatory cognitive robots,'' \emph{{arXiv}}, 2024.

\bibitem{dennis2016practical}
L.~A. Dennis, M.~Fisher, N.~K. Lincoln, A.~Lisitsa, and S.~M. Veres, ``Practical verification of decision-making in agent-based autonomous systems,'' \emph{Automated Software Engineering}, vol.~23, no.~3, pp. 305--359, 2016.

\bibitem{scheutz2013novel}
M.~Scheutz, G.~Briggs, R.~Cantrell, E.~Krause, T.~Williams, and R.~Veale, ``Novel mechanisms for natural human-robot interactions in the diarc architecture,'' in \emph{Proceedings of AAAI workshop on intelligent robotic systems}.\hskip 1em plus 0.5em minus 0.4em\relax Palo Alto, CA, 2013, p.~66.

\bibitem{schermerhorn2006diarc}
P.~W. Schermerhorn, J.~F. Kramer, C.~Middendorff, and M.~Scheutz, ``Diarc: A testbed for natural human-robot interaction.'' in \emph{AAAI}, vol.~6, 2006, pp. 1972--1973.

\bibitem{english2020ontoagent-87e}
J.~English and S.~Nirenburg, ``{OntoAgent}: Implementing content-centric cognitive models,'' in \emph{Proceedings of the Annual Conference on Advances in Cognitive Systems}, ser. Eighth Annual Conference on Advances in Cognitive Systems, 2020.

\bibitem{mcshane2021linguistics-995}
M.~{McShane} and S.~Nirenburg, \emph{Linguistics for the Age of {AI}}.\hskip 1em plus 0.5em minus 0.4em\relax Cambridge, {MA}: {MIT} Press, 2021.

\bibitem{nirenburg2024mutual-5e0}
S.~Nirenburg, T.~Ferguson, and M.~{McShane}, ``Mutual trust in human- ai teams relies on metacognition,'' in \emph{Metacognitive Artificial Intelligence}, Wei, {\textbackslash}.~{Shakarian, Hui and{\textbackslash} }, and Paulo, Eds.\hskip 1em plus 0.5em minus 0.4em\relax Cambridge University Press, 2024.

\bibitem{colledanchise2018behavior}
M.~Colledanchise and P.~{\"O}gren, \emph{Behavior Trees in Robotics and AI: An Introduction}.\hskip 1em plus 0.5em minus 0.4em\relax CRC Press, 2018.

\bibitem{iannotta2022heterogeneous}
M.~Iannotta, D.~C. Dom{\'\i}nguez, J.~A. Stork, E.~Schaffernicht, and T.~Stoyanov, ``Heterogeneous full-body control of a mobile manipulator with behavior trees,'' \emph{arXiv preprint arXiv:2210.08600}, 2022.

\bibitem{oruganti2025ikt1}
S.~S.~O. Venkata, R.~Parasuraman, and R.~Pidaparti, ``Ikt-bt: Indirect knowledge transfer behavior tree framework for multirobot systems through communication eavesdropping,'' \emph{IEEE Transactions on Cybernetics}, 2025.

\bibitem{venkata2023kt}
------, ``Kt-bt: A framework for knowledge transfer through behavior trees in multirobot systems,'' \emph{IEEE Transactions on Robotics}, vol.~39, no.~5, pp. 4114--4130, 2023.

\bibitem{colledanchise2021implementation}
M.~Colledanchise and L.~Natale, ``On the implementation of behavior trees in robotics,'' \emph{IEEE Robotics and Automation Letters}, vol.~6, no.~3, pp. 5929--5936, 2021.

\bibitem{kahneman2011thinking}
D.~Kahneman, \emph{Thinking, Fast and Slow}.\hskip 1em plus 0.5em minus 0.4em\relax Macmillan, 2011.

\bibitem{laird2012cognitive}
\BIBentryALTinterwordspacing
J.~E. Laird, \emph{The Soar Cognitive Architecture}.\hskip 1em plus 0.5em minus 0.4em\relax Cambridge, MA: MIT Press, 2012. [Online]. Available: \url{https://mitpress.mit.edu/9780262122962/the-soar-cognitive-architecture/}
\BIBentrySTDinterwordspacing

\bibitem{ritter2019act}
\BIBentryALTinterwordspacing
F.~E. Ritter, J.~R. Anderson \emph{et~al.}, ``Act-r: A cognitive architecture for modeling cognition,'' \emph{Wiley Interdisciplinary Reviews: Cognitive Science}, vol.~10, no.~3, p. e1488, 2019. [Online]. Available: \url{https://doi.org/10.1002/wcs.1488}
\BIBentrySTDinterwordspacing

\bibitem{scheutz2019overview}
M.~Scheutz, T.~Williams, E.~Krause, B.~Oosterveld, V.~Sarathy, and T.~Frasca, ``An overview of the distributed integrated cognition affect and reflection diarc architecture,'' \emph{Intelligent Systems, Control and Automation: Science and Engineering}, pp. 165--193, 2019.

\bibitem{laird1991robosoar}
\BIBentryALTinterwordspacing
J.~E. Laird, P.~S. Rosenbloom, and A.~Newell, ``Robo-soar: An integration of external interaction, planning, and learning using soar,'' \emph{Robotics and Autonomous Systems}, vol.~8, no. 1-2, pp. 113--129, 1991. [Online]. Available: \url{https://doi.org/10.1016/0921-8890(91)90017-F}
\BIBentrySTDinterwordspacing

\bibitem{puigbo2015gpsr}
\BIBentryALTinterwordspacing
J.-Y. Puigbo, A.~Pumarola, C.~Angulo, and R.~Tellez, ``Using a cognitive architecture for general purpose service robot control,'' \emph{Connection Science}, vol.~27, no.~2, pp. 105--117, 2015. [Online]. Available: \url{https://doi.org/10.1080/09540091.2014.968093}
\BIBentrySTDinterwordspacing

\bibitem{harnad1990symbol}
\BIBentryALTinterwordspacing
S.~Harnad, ``The symbol grounding problem,'' \emph{Physica D: Nonlinear Phenomena}, vol.~42, no. 1-3, pp. 335--346, 1990. [Online]. Available: \url{https://doi.org/10.1016/0167-2789(90)90087-6}
\BIBentrySTDinterwordspacing

\bibitem{trafton2013actre}
\BIBentryALTinterwordspacing
J.~G. Trafton, L.~M. Hiatt, A.~M. Harrison, I.~Franklin P.~Tamborello, S.~S. Khemlani, and A.~C. Schultz, ``Act-r/e: An embodied cognitive architecture for human–robot interaction,'' \emph{Journal of Human-Robot Interaction}, vol.~2, no.~1, pp. 30--55, 2013. [Online]. Available: \url{https://doi.org/10.5898/JHRI.2.1.Trafton}
\BIBentrySTDinterwordspacing

\bibitem{scheutz2011humanlike}
M.~Scheutz, R.~Cantrell, and P.~Schermerhorn, ``Toward humanlike task-based dialogue processing for human robot interaction,'' \emph{AI Magazine}, vol.~32, no.~4, pp. 77--84, 2011.

\bibitem{oruganti2023automating}
S.~Oruganti, S.~Nirenburg, J.~English, and M.~McShane, ``Automating knowledge acquisition for content-centric cognitive agents using llms,'' in \emph{Proceedings of the AAAI Symposium Series}, vol.~2, no.~1, 2023, pp. 379--385.

\bibitem{nirenburg2023hybrid}
S.~Nirenburg, N.~Krishnaswamy, and M.~McShane, ``Hybrid ml/kb systems learning through nl dialog with dl models,'' in \emph{AAAI-Make Workshop on Challenges Requiring the Combination of Machine Learning and Knowledge Engineering}, 2023.

\bibitem{nirenburg2021overcoming}
S.~Nirenburg, M.~McShane, and J.~English, ``Overcoming the knowledge bottleneck using lifelong learning by social agents,'' in \emph{International Conference on Applications of Natural Language to Information Systems}.\hskip 1em plus 0.5em minus 0.4em\relax Springer, 2021, pp. 24--29.

\end{thebibliography}

\end{document}